\documentclass{article}

\usepackage[preprint]{neurips_2026}

\usepackage[utf8]{inputenc} 
\usepackage[T1]{fontenc}    
\usepackage{hyperref}       
\usepackage{url}            
\usepackage{booktabs}       
\usepackage{amsfonts}       
\usepackage{nicefrac}       
\usepackage{microtype}      
\usepackage{xcolor}         
\usepackage{algorithm}
\usepackage{algpseudocode}
\usepackage{float}
\usepackage{enumitem}
\usepackage{graphicx}
\usepackage{setspace}
\usepackage{colortbl}
\usepackage{booktabs}
\usepackage{amsmath}
\usepackage{natbib}

\definecolor{mementobg}{RGB}{232, 245, 240}
\definecolor{deltagreen}{RGB}{15, 110, 86}
\definecolor{sectionbg}{RGB}{242, 242, 245}
\definecolor{bestbg}{RGB}{215, 235, 255}
\definecolor{besttext}{RGB}{12, 68, 124}
\definecolor{aetbg}{RGB}{255, 245, 220}
\definecolor{aettext}{RGB}{133, 79, 11}
\definecolor{deltared}{RGB}{163, 45, 45}

\usepackage[capitalize,noabbrev]{cleveref}

\usepackage{graphicx}

\usepackage{fancyhdr}
\renewcommand{\headrulewidth}{0.4pt}

\title{MEMENTO: Leveraging Web as a Learning Signal for Low-Data Domains}

\author{
\textbf{Ashutosh Ojha} \hspace{0.5em}
\textbf{Vinay Aggarwal} \hspace{0.5em}
\textbf{Ashutosh Srivastava} \\
\textbf{Siddharth Yedlapati} \hspace{0.5em}
\textbf{Yaman K Singla} \hspace{0.5em}
\textbf{Jitendra Ajmera} \\
\vspace{0.4em}
\raisebox{-0.20em}{\includegraphics[height=1.6em]{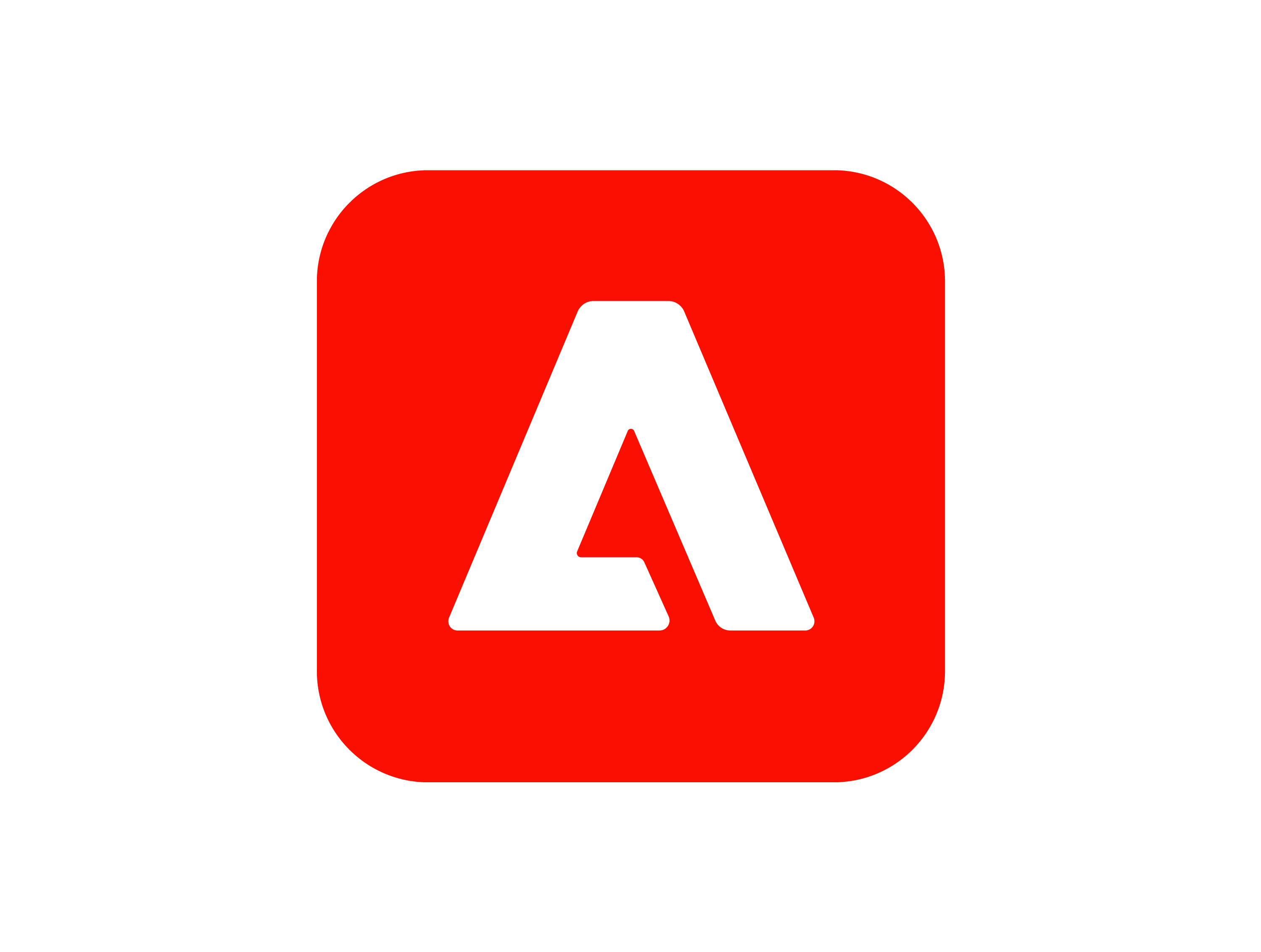}}%
\hspace{0.15em}%
Adobe, Media \& Data Science Research \\
\vspace{-0.5em}
\texttt{behavior-in-the-wild@googlegroups.com}
}

\begin{document}

\maketitle

\begin{abstract}

Real-world tasks often lack large labeled datasets, motivating extensive work on learning in low-data regimes. Existing approaches such as few-shot prompting, instruction tuning, and synthetic data generation, continue to treat labeled or pseudo-labeled data as the primary learning signal. In contrast, human practitioners acquire expertise through repeated, self-directed interaction with the open web, progressively refining both domain knowledge and search strategies. We propose \textbf{MEMENTO}, a framework that treats the web as a learning signal rather than a stateless retrieval interface. MEMENTO operates at two levels: within each session, it conducts iterative web exploration via an \emph{Adaptive Exploration Tree} (AET) that decomposes tasks into evolving questions and reflects on intermediate findings; across sessions, it accumulates experience through dual-channel memory, separating declarative knowledge (facts) from procedural knowledge (search strategies). This design enables agents to learn reusable research strategies and domain expertise from trajectories of web interaction without additional model training. We evaluate MEMENTO on three structurally distinct low-data domains: Sales Automation, Legal Outcome Prediction, and Subpopulation Opinion Prediction. Our empirical results show consistent improvement in performance over ReAct baselines ($+25.6\%$ on sales, $+36.5\%$ on legal research, and $29.6\%$ TVD reduction on opinion prediction), demonstrating that the web can serve as a scalable learning source for acquiring task-specific expertise in data-scarce settings.
\end{abstract}


\section{Introduction}

Large labeled datasets are available for only a handful of well-studied tasks, whereas the majority of real-world tasks suffer from severe data scarcity. Domains such as legal research, sales strategy, and medical diagnosis are highly context-specific, and expert annotation at scale is infeasible~\citep{brown2020language, hedderich2021survey, chalkidis2022lexglue, rajpurkar2022ai}. This data scarcity has attracted sustained research attention on learning in low-data regimes.

Prior work has approached this problem primarily through better use of labeled data or model pretraining. T5~\citep{raffel2020exploring} showed that multi-task pretraining improves generalization under limited supervision. GPT-3~\citep{brown2020language} demonstrated few-shot adaptation from in-context examples. Instruction finetuning~\citep{wei2021finetuned} improved zero-shot generalization across diverse task collections, and synthetic data generation~\citep{wang2023self, gunasekar2023textbooks} extended adaptation to settings with minimal annotations. Despite these advances, all of these approaches treat labeled or pseudo-labeled data as the primary learning signal.

We propose the \emph{open web} as an alternative learning signal for low-data settings. In domains where labeled supervision is scarce, human practitioners do not wait for thousands of annotated examples. They learn through repeated, self-directed engagement with web resources. A junior lawyer assigned a novel case consults legal databases, follows citation trails, and reads practitioner forums, gradually developing a sense of reliable sources and strategies. A sales representative entering a new market reads industry publications, studies competitor analyses, and internalizes which signals indicate genuine buyer intent. The web is thus uniquely suited to low-data regimes: it is openly accessible, requires no annotation, and carries both factual content and implicit domain expertise.

Current LLM agents do not fully leverage this. Most treat web search as a stateless retrieval tool~\citep{nakano2021webgpt, openai2025deepresearch}: issuing queries, extracting facts, and discarding the experience accumulated along the way. Beyond direct factual content, the web contains expert-authored material such as practitioner forums, domain blogs, and methodology articles that encodes how experienced practitioners approach problem classes. Current agents have no mechanism to learn from such content across sessions, leaving the web’s value as a source of transferable domain expertise entirely untapped.

We introduce \textbf{MEMENTO}, a framework that uses the web as a learning signal for craft acquisition in low-data settings. MEMENTO operates at two levels. Within each session, an \textbf{Adaptive
Exploration Tree} (AET) conducts research much like a lawyer working a
novel case: AET decomposes the research task into an initial set of questions and uses web search to answer them. The agent then reflects on all responses, identifies remaining gaps, and generates a new set of questions, repeating this cycle under a fixed budget. Information discovered at any point is stored in a local session memory, ensuring that each reflective cycle builds on what has already been found.


Across sessions, MEMENTO maintains two persistent and independent memory systems grounded in the declarative/procedural distinction from the integrated theory of mind~\citep{anderson2004integrated}. A \textbf{Declarative Memory} stores source-grounded facts and context from prior sessions, allowing future tasks to build on established foundations rather than starting from scratch. A \textbf{Procedural Memory} distills high-utility query strategies, source credibility heuristics, and domain-specific research playbooks from prior trajectories. Together, these components form an iterative learning pipeline: cross-session memory loads at the start of each session to initialize the AET with accumulated strategy and knowledge; the AET generates a rich trajectory through directed, reflective search; and that trajectory updates cross-session memory at the session's end. MEMENTO thus separately tracks what it has learned and how it learned it, refining both across repeated web engagements.

We evaluate MEMENTO on three low-data domains with different output structures: \textbf{Sales Automation} (synthesizing evidence-backed value propositions), \textbf{Legal Research} (predicting outcomes from neutral case facts), and \textbf{Subpopulation Opinion Prediction} (estimating a demographic subgroup's response distribution). MEMENTO's consistent gains across three tasks suggest the improvements stem from the cross-session learning architecture rather than a domain-specific effect. We make the following contributions:

\begin{itemize}[leftmargin=*]

\item \textbf{Web as a Learning Signal in Low-Data Regimes:} To the best of our knowledge, we are the first to propose the open web as a learning signal for craft acquisition in domains where labeled supervision is scarce. MEMENTO uses research trajectories and accumulated factual knowledge to perform cross-session optimization, enabling agents to develop domain expertise without model fine-tuning.

\item \textbf{Adaptive Exploration Tree:} We introduce a within-session search architecture that decomposes research tasks into dynamic question trees, iteratively reflecting on discovered information to identify gaps and restructure search paths under a fixed budget, producing trajectories that are directed, self-correcting, and informative as cross-session learning signals.

 \item \textbf{Dual-Channel Cross-Session Memory:} Grounded in the declarative/procedural distinction from the integrated theory of mind~\citep{anderson2004integrated}, MEMENTO maintains two independent persistent memory stores. Procedural Memory distills query strategies, source credibility heuristics, and domain-specific search playbooks from prior trajectories; Declarative Memory persists source-grounded facts and contextual knowledge across sessions, allowing research strategy and domain knowledge to develop independently.
 
\item \textbf{Empirical Validation:} We evaluate MEMENTO on three domains: Sales Automation, Legal Research and Subpopulation Opinion Prediction -- spanning free-form generation, classification, and distribution prediction. MEMENTO demonstrates consistent improvement over ReAct baselines (+$25.6\%$ on Sales Automation, +$36.5\%$ on Legal Research, and $29.6\%$ TVD reduction on Opinion Prediction), with gains that persist against three enhanced retrieval baselines (query rewriting, source re-ranking, and their combination) and on a post-cutoff legal test set constructed to rule out pretraining exposure.

\end{itemize}

\section{Related Work}

\begin{table}[t]
\centering
\scriptsize
\resizebox{\linewidth}{!}{
\begin{tabular}{lcccll}
\toprule
\textbf{System} & \textbf{Live web} & \textbf{Cross-session} & \textbf{Weight updates} & \textbf{Persistent artifact} & \textbf{Primary target} \\
\midrule
ReAct / Deep Research & Yes & No & No & None & Single-session research \\
DeepResearcher & Yes & No & Yes & Search policy & Open-domain question answering \\
WebCoach & Yes & Yes & No & Episodic navigation traces & Browser navigation \\
MemoryR1 & N/A & Yes & Yes & Managed conversational facts & Long-horizon dialogue \\
\textbf{MEMENTO} & Yes & Yes & No & \textbf{Procedures + Domain Facts} & \textbf{Low-data Craft Acquisition} \\
\bottomrule
\end{tabular}}
\caption{Comparison of MEMENTO with related works. \emph{Cross-session}: the system retains knowledge from prior tasks that improves performance on future tasks.}
\label{tab:positioning}
\end{table}

\textbf{Learning in Low-Data Regimes.} Few-shot learning~\citep{brown2020language}, instruction finetuning~\citep{wei2021finetuned}, in-context learning~\citep{dong2024survey}, and synthetic data generation~\citep{wang2023self, gunasekar2023textbooks} have each pushed the boundary of what is possible with scarce supervision. Despite their diversity, all share a common assumption: learning is bounded by a fixed training signal, whether labeled examples, instructions, or synthetic data. These approaches have no mechanism to actively seek out new information when supervision is scarce. MEMENTO instead treats the open web as an additional cross-session signal: unsupervised web trajectories update craft and search knowledge, while a small labeled set supervises decomposition-rule refinement, all without model fine-tuning.

\textbf{Web as a Tool and as Supervision.} A growing body of work equips LLMs with web access at varying levels of sophistication. RAG~\citep{lewis2020retrieval} and its extensions~\citep{trivedi2023interleaving, schick2023toolformer, fan2024survey} ground outputs in retrieved context. WebGPT~\citep{nakano2021webgpt}, STORM-style research systems~\citep{shao-etal-2024-assisting,jiang-etal-2024-unknown}, and Deep Research agents~\citep{openai2025deepresearch, google2024deepresearch, perplexity2025deepresearch, xai2025grok3,zheng2025deepresearcher} add active browsing, planning, and multi-step synthesis. These systems primarily optimize the current episode: retrieved evidence is consumed as context, not distilled into separately evolving factual and procedural stores. Web-supervised representation learning, pseudo-labeling, and synthetic-data methods likewise convert web content into a fixed training set. MEMENTO instead preserves the \emph{experience of retrieval itself} as a cross-session learning signal, allowing search strategy and domain knowledge to improve independently.

\textbf{Within-Session Search Architectures.} ReAct~\citep{yao2022react} interleaves reasoning and action in linear sequences, while Tree of Thoughts~\citep{yao2023tree} and related approaches introduce branching over candidate reasoning paths. These methods improve over single-pass retrieval but remain limited in their ability to dynamically restructure search based on what is discovered mid-session. MEMENTO's Adaptive Exploration Tree (AET) addresses this by decomposing tasks into dynamic question trees that continuously reflect on accumulated findings, identify remaining gaps, and restructure search paths under a fixed budget. Unlike static branching approaches, the AET treats mid-session discoveries as explicit signals for restructuring subsequent search rather than as passive context, producing trajectories that are directed and self-correcting rather than exhaustive.

\textbf{RL and Memory-Augmented Agents.} A growing body of work applies reinforcement learning to train LLM agents for retrieval tasks~\citep{wei2025webagent, chen2025learning, song2025r1, jin2025search}, with systems such as MemoryR1~\citep{yan2025memory}, SE-Search~\citep{li2026se}, and WebCoach~\citep{liu2025webcoach} introducing explicit memory or cross-session guidance. These results establish that retrieval policies and persistence matter. MEMENTO differs in three respects: it requires no weight updates or large-scale RL rollouts; it explicitly separates procedural from declarative memory; and it audits how rules are supported, contradicted, merged, and pruned across batches. This design targets labeled-data scarcity while retaining interpretable memory artifacts that can be inspected or removed.

\textbf{Learning from Data at the Prompt Level.} Recent works including GEPA~\citep{agrawal2025gepa} and MIPROv2~\citep{opsahl2024optimizing} optimize LLM behavior at the prompt level without modifying weights. \citet{agrawal2025gepa} reflects on execution trajectories in natural language to iteratively evolve system prompts, while \citet{opsahl2024optimizing} jointly tunes instructions and few-shot demonstrations via Bayesian optimization. However, both are fundamentally dataset-driven: the learning signal comes entirely from labeled examples, meaning the quality of the learned artifact is bounded by the coverage and size of the training set. MEMENTO takes a different approach entirely; rather than learning from a fixed labeled dataset, it uses the open web as its primary training signal, allowing craft knowledge and domain expertise to be acquired primarily from web interaction.


Table~\ref{tab:positioning} summarizes how MEMENTO's combination of live web, cross-session learning, memory modules, and no weight updates distinguishes it from the closest systems.

\section{Methodology}

\subsection{Problem Formulation}
We consider tasks of the form: given an input context $x$ (a question, case, scenario, or specification), produce an output $\hat{y}$ (a prediction, analysis, or generated artifact) that can be evaluated against a ground truth $y$. We assume a training corpus
\[
\mathcal{D} = \{(x_i, y_i)\}_{i=1}^N
\]
of examples, and a held-out test set $\mathcal{D}_{\mathrm{test}}$. The agent has access to a web search tool with a date cutoff enforced at both training and inference time, preventing it from retrieving information about the outcome it is trying to predict. The goal is not to fine-tune model weights but to learn a set of transferable reasoning artifacts, namely procedural strategies and declarative facts, that accumulate across training examples and improve the agent's performance and efficiency on future unseen inputs.

\subsection{System Overview}

\begin{figure}[htbp]
    \centering
    \includegraphics[width=\textwidth]{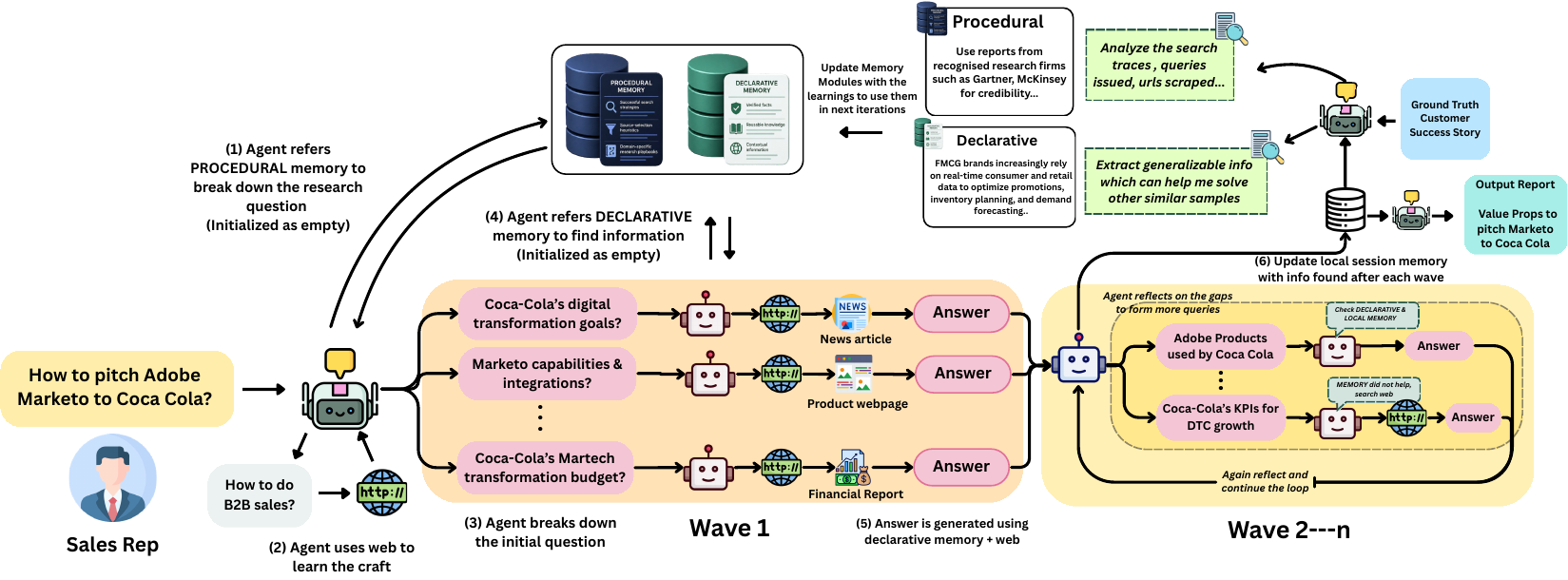}
    \caption{Overview of MEMENTO for a single training sample. Given a research 
question, the agent (1)~consults procedural memory for question decomposition or (2)~optionally explores the web to bootstrap craft knowledge 
when memory is empty. (3)~The root is decomposed into Wave~1 sub-questions, 
each question is solved by a tool-augmented agent that (4)~checks declarative memory 
before issuing date-restricted web searches and (5)~returns an answer. In 
subsequent waves, reflection identifies gaps and adds new sub-questions, 
prioritizing memory over web retrieval. (6)~Local session memory is updated 
after each wave; at the end of the sample, the trajectory is distilled into 
procedural rules and declarative facts that update the persistent memory 
stores.}
    \label{fig:pipeline}
\end{figure}

Figure~\ref{fig:pipeline} provides an overview of the MEMENTO framework. MEMENTO operates in two interleaved phases that repeat over batches of training examples. In the \textbf{forward pass}, each input $x$ is processed by the Adaptive Exploration Tree (AET), a hierarchical decomposition and search architecture that breaks the task into sub-questions, answers each through web interaction, and synthesizes a final output $\hat{y}$. The AET is initialized at the start of each session with the current state of two persistent memory stores, Procedural Memory and Declarative Memory, which provide accumulated search strategies and domain knowledge from prior batches. In the \textbf{batch update}, after all examples in a batch are processed, the system reflects on execution trajectories and ground-truth outcomes to update both memory stores, which are then carried forward to the next batch. Together, these two phases form a closed learning loop: the forward pass generates rich research trajectories that the batch update distills into reusable expertise, and that expertise in turn improves future forward passes. At inference time, only the forward pass runs with the final learned memory state frozen and no further updates occur.

\subsection{Within-Session Research: The Adaptive Exploration Tree}

Rather than asking the LLM to directly produce $\hat{y}$ from $x$, MEMENTO frames $x$ as a 
root question and decomposes it into a tree of sub-questions $Q = \{q_0, q_1, \ldots, q_k\}$, where $q_0$ is the root. The Adaptive Exploration Tree (AET) governs within-session research through three tightly coupled components: wave-based decomposition, which builds the tree adaptively; tool-augmented agents, which resolve individual sub-questions; and bottom-up synthesis, which aggregates answers into a final output. Throughout, a local session memory accumulates findings incrementally, ensuring that every component sees what has already been discovered.

\paragraph{Local session memory.}
AET maintains a local session memory that grows as sub-questions are resolved. Each solved sub-question contributes its answer and supporting evidence to this memory. This memory is available to all components of the AET: the reflection agent consults it when generating new waves of sub-questions, and individual tool-augmented agents query it before issuing external web searches. At the end of the forward pass, this memory contains the full research trajectory for the sample, which is used at the end of each training batch to update the persistent cross-session memory stores (Section~\ref{sec:persistent_memory}).

\paragraph{Wave-based decomposition.}
AET decomposes $x$ in waves. The first wave breaks the root question into an initial set of sub-questions up to a configurable budget $B_0$. Between waves, a reflection agent is invoked: it receives the answers generated so far, intermediate findings, and prior reflection history, and produces a new set of sub-questions for the next wave conditioned on what has already been learned -- targeting dependencies, ambiguities, or overlooked dimensions surfaced by earlier answers. This process repeats until the overall question budget is exhausted. Since each wave is informed by prior results, the tree is iteratively reshaped rather than statically defined.

\paragraph{Sub-question solving: tool-augmented agents.}
Each sub-question $q_j$ is assigned to an independent tool-augmented agent with four primitive actions: \textsc{Search\_Memory} (retrieves info from local session memory and the long-term declarative memory store before incurring external retrieval), \textsc{Search\_Web} (issues a web query returning a ranked list of results), \textsc{Scrape\_Results} (fetches full textual content of selected results, cached to avoid redundant fetches) and \textsc{Generate\_Answer} (synthesizes the accumulated evidence into a final response).

\paragraph{Bottom-up synthesis.}
Once all waves are complete, answers are synthesized upward through the tree. Only nodes decomposed during reflection (having children but no direct answer) require synthesis. Each such node receives the answers of its children and produces a synthesized answer via an LLM call, proceeding from the deepest level to the shallowest. The root node's synthesized answer is the final output $\hat{y}$.

\subsection{Persistent Memory Architecture}
\label{sec:persistent_memory}
Across batches, MEMENTO maintains two memory systems grounded in the procedural-declarative distinction from the integrated theory of mind~\cite{anderson2004integrated}: procedural memory captures knowledge of \textit{how} to research, and declarative memory captures knowledge of
\textit{what} is known about the domain. The two systems are independent in scope but share a batch-level update cadence (Section~\ref{sec:batch_update}). Please refer Appendix~\ref{app:memory-examples} to see qualitative examples of each of these memories.

\paragraph{Procedural Memory: three complementary stores.}
Procedural memory is split into three stores that target different stages of the AET pipeline.

\textbf{Craft Knowledge ($M_1$)} is a free-text store of high-level research strategy, which sub-questions tend to be informative, what sources are authoritative, how to reason under uncertainty etc. $M_1$ is learned \emph{entirely from the web, without supervision}: each sample's local session memory (a running synthesis of web evidence) is distilled by an LLM into generalizable insights and aggregated into $M_1$ at the end of every batch. No ground truth is consulted; the supervisory signal is the web evidence itself. $M_1$ is consumed at the \emph{planning} stage (by the wave-zero decomposer and the reflection agent) and is the only store that supports \emph{cold-start enrichment}. Before any web search, the agent runs a relevance filter over $M_1$ to extract applicable craft knowledge to the current sample. Agent issues exploratory searches only if no sufficient match is found distilling the traces into new craft entries.

\textbf{Decomposition Rules ($M_2$)} is a structured catalog of
conditional rewrite rules of the form ``\textsc{When} \textless
pattern in root or sample inputs\textgreater, \textsc{add} a
sub-question that \textsc{covers} \textless pattern\textgreater.''
While $M_1$ encodes strategic preferences in free-form prose, $M_2$
encodes hard, sample-conditioned compositional patterns. $M_2$ is
consumed at the \emph{decomposition} stage: rules are injected at
every depth of the question-breakdown prompt, both at wave zero and during reflection.


\textbf{Web Action Rules ($M_3$)} is a structured catalog of rules governing the \emph{tool-use} stage: query formulation (e.g., quoted-phrase length, language separation), URL selection (e.g., authority filters, $\textit{site:}$ qualifiers), and scraping constraints. $M_3$ is injected into the system prompt of every tool-augmented agent, so it
shapes execution \emph{inside} a sub-question solve rather than the tree structure itself. Each rule carries a \texttt{support\_count} and a
\texttt{contradiction\_count} across batches; rules are strengthened when independent trajectories reproduce them and down-weighted when contradictions exceed support.

\paragraph{Declarative Memory} comprises a single \textbf{Domain Store ($M_4$)}: a structured free-text repository of entity-specific factual knowledge extracted from prior samples. At inference time, a retrieval step retrieves the current sample's declarative cache against $M_4$'s keys and injects the matching subset into the solver as always-visible context.

\subsection{Cross-Session Learning: The Batch Update Loop}
\label{sec:batch_update}

\textbf{Phase A — Forward Pass and Unsupervised Consolidation.} All $b$ examples are processed in parallel using the current memory state $\mathcal{M} = \langle M_1, M_2, M_3, M_4 \rangle$. Post-inference, two unsupervised consolidation steps run without access to ground truth: (i) craft knowledge $M_1$ is rewritten by an LLM aggregating per-sample methodology learnings from valid runs, and (ii) search rules $M_3$ are updated from search execution logs. Domain store $M_4$ is then updated by folding entity-specific facts extracted during the runs. Before storage, candidates pass a correctness/sufficiency check (including removal of ``no data'' outputs) and a canonicalization pass that merges near-duplicates. For $M_3$, independently recurring or conflicting batch evidence updates each rule's support and contradiction counts.

\textbf{Phase B — Scoring.} Each predicted output $\hat{y}_i$ is scored against its ground truth $y_i^*$ using an LLM judge, producing a normalized score per sample. Failed runs are excluded from the subsequent supervised phase.

\textbf{Phase C — Supervised Reflection.} For each successfully scored run, the system compares predicted outputs against ground truth at a fine-grained level. For each target element, the system inspects the question decomposition tree to determine whether the corresponding sub-question was present and whether it led to correct coverage. Rules active in successful trajectories accumulate \textit{support}; rules present in failed trajectories accumulate \textit{contradiction}. An LLM proposes additions, reinforcements, or removals, which are canonicalized and merged into $M_2$ according to their support-to-contradiction ratio across batches. The updated $\mathcal{M}$ is carried forward to the next batch. Appendix~\ref{app:memory-audit} reports how these statistics and memory size evolve over all six training batches.

At test time, the forward pass runs with frozen memory. No updates occur. All memory stores are injected as context, and the system produces $\hat{y}$ using the same pipeline as training. The learned knowledge remains interpretable as human-readable text.



   

\section{Experimental Setup}

\subsection{Tasks and Datasets}
\label{sec:tasks-and-datasets}

\paragraph{Task 1: Sales Automation.}
Given a triple of (seller company, product list, prospective customer), the system predicts the value propositions a sales representative would use to pitch the given products to the given customer, each capturing a pain point and the mechanism by which the product addresses it. The agent receives only the input triple and must recover these propositions entirely through open-web research, mirroring the workflow of a Sales Development Representative synthesizing publicly available information into an evidence-backed pitch. We use the SDR-Bench dataset~\cite{sdrbench2026}, consisting of publicly available customer success stories documenting prior deals between enterprises. The dataset comprises 180 seller-customer-product triples, each paired with a ground truth pitch comprising value propositions and supporting evidence. To prevent temporal leakage, web search is restricted to sources published more than six months before each story's publication date. The samples are partitioned with seed 42 into 60 training and 120 test samples.

\paragraph{Task 2: Legal Research.}
Given the neutral facts of a U.S. Supreme Court case, the system predicts the judgment outcome: the winning party, the disposition of the court's decision, and a reasoned conclusion. We use the JUSTICE benchmark~\citep{alali2021justice}, a dataset of U.S. Supreme Court cases sourced from Oyez. We restrict to cases decided after 2010, yielding 722 candidates with sufficient web coverage. To prevent the agent from trivially retrieving the exact case record, we apply a leakage filtering step: for each candidate, a web agent is given the case facts with a search cutoff of five years prior to the decision date; cases where the agent successfully retrieves the exact outcome are discarded. After filtering, we take 180 samples partitioned with seed 42 into 60 training and 120 test samples. The five-year backdating is enforced consistently at both training and inference time. Since JUSTICE cases
were compiled by 2021 and may appear in pretraining corpora of the backbone models, we additionally construct a 41-case test set from the 2025--2026 Supreme Court term to evaluate post-cutoff performance (Appendix~\ref{app:recent-legal}).

\paragraph{Task 3: Subpopulation Opinion Prediction.}
Given a survey question and a target subgroup, the system predicts the subgroup's full probability distribution over the fixed answer options. We use SubPOP~\citep{subpop2025}, drawn from Pew Research Center's American Trends Panel~\citep{pew_atp} and NORC's General Social Survey~\citep{gss2022}, and evaluate on 100 questions spanning eight demographic attributes: region, education, income, ideology, political party, race, religion, and sex. The task is low-data in the same sense as the other two: subgroup-level ground truth exists only for questions that have already been fielded, and the relevant evidence (related polls, demographic research, issue coverage) is scattered across heterogeneous web sources that must be located and combined. Because the output is a distribution rather than free-form text or a binary label, this task tests whether learned research procedures transfer to a third output structure. Full setup and results are in Appendix~\ref{app:subpop}.

\subsection{Baselines}

We compare \textbf{MEMENTO} against (i) \emph{closed-book LLMs}, (ii) \emph{five-shot in-context learning}, (iii) \emph{ReAct with web search}, (iv) \emph{ReAct with MEMENTO memory injected} (isolating memory from AET), (v) \emph{AET without memory} (isolating within-session search), and (vi) three enhanced \emph{ReAct variants with query rewriting, reranking, and their combination} for a controlled sales-task re-run (Appendix~\ref{app:strong-baselines}).


\subsection{Evaluation Metrics}

\paragraph{Sales Automation.}
Predictions are evaluated by an LLM judge prompted to act as a Senior Sales Enablement Evaluator. We extract the ground truth value propositions from the actual customer success story ($y$) and the synthesized value propositions from the report generated by MEMENTO ($\hat{y}$). For each sample, we employ the judge to score if each ground truth value proposition is covered in the generated output or not. Each GT value prop is scored on a 0--5 sales-effectiveness scale, ranging from a complete miss (0) to full recovery of the product, pain point, mechanism, and supporting evidence (5). Full rubric details are provided in Appendix~\ref{app:rubric}. We report the \textbf{average coverage score}, the mean of coverage scores across all the samples in the test set.

\paragraph{Legal Research.}
We evaluate this as a binary win-prediction problem. For each case, the system outputs a predicted winning party (petitioner or respondent), which is scored against the ground-truth \texttt{winning\_party} field from JUSTICE using an LLM judge returning a binary \textsc{correct}/\textsc{incorrect} verdict. The reported metric is the prediction accuracy.

\paragraph{Subpopulation Opinion Prediction.}
\label{sec:subpop-metrics}
We report Total Variation Distance (TVD) and KL divergence between the predicted and observed subgroup distributions; lower is better for both (definitions in Appendix~\ref{app:subpop}).

\subsection{Implementation Details}
We use two backbone LLMs: Qwen~\citep{yang2025qwen3}, an open model, and GPT-5-mini~\citep{openai2025gpt5mini}, a frontier model, for cross-model validation. AET operates under three configurable budgets: total sub-questions, per-question agent steps, and cold-start craft exploration. Training proceeds in batches of $b$ examples. Sales outputs are scored by Qwen-3.5 as the primary judge; we additionally validate the evaluation using GPT-4o and GPT-5.4 as independent judges, with inter-judge agreement reported in Appendix~\ref{app:judge-agreement}. Full hyperparameters are in Appendix~\ref{app:hyperparams}.

\begin{table}[t]
\centering
\small
\setlength{\tabcolsep}{4pt}
\renewcommand{\arraystretch}{0.9}
\begin{tabular}{l cc}
\toprule
& \shortstack{\textbf{Sales Automation} \\ \textit{(Coverage Score)}} 
& \shortstack{\textbf{Legal Research} \\ \textit{(Accuracy)}} \\
\midrule
\rowcolor{sectionbg}
\multicolumn{3}{l}{\textbf{LLM (only parametric knowledge)}} \\
Qwen & 0.416 & 0.808 \\
\rowcolor{mementobg}
Qwen + \textsc{Memento} & 0.437 \textcolor{deltagreen}{\scriptsize(+0.021)} & 0.808 \textcolor{deltagreen}{\scriptsize(+0.000)} \\
GPT-5-mini & 0.391 & 0.700 \\
\rowcolor{mementobg}
GPT-5-mini + \textsc{Memento} & 0.398 \textcolor{deltagreen}{\scriptsize(+0.007)} & 0.717 \textcolor{deltagreen}{\scriptsize(+0.017)} \\
\midrule
\rowcolor{sectionbg}
\multicolumn{3}{l}{\textbf{In-context learning}} \\
5-shot Qwen & 0.375 & 0.842 \\
\rowcolor{mementobg}
5-shot Qwen + \textsc{Memento} & 0.382 \textcolor{deltagreen}{\scriptsize(+0.007)} & 0.852 \textcolor{deltagreen}{\scriptsize(+0.010)} \\
5-shot GPT-5-mini & 0.462 & 0.692 \\
\rowcolor{mementobg}
5-shot GPT-5-mini + \textsc{Memento} & 0.476 \textcolor{deltagreen}{\scriptsize(+0.014)} & 0.675 \textcolor{deltared}{\scriptsize(-0.017)} \\
\midrule
\rowcolor{sectionbg}
\multicolumn{3}{l}{\textbf{Web as a tool (ReAct)}} \\
ReAct + Qwen & 0.461 & 0.592 \\
\rowcolor{mementobg}
ReAct + Qwen + \textsc{Memento} & 0.469 \textcolor{deltagreen}{\scriptsize(+0.008)} & 0.742 \textcolor{deltagreen}{\scriptsize(+0.150)} \\
ReAct + GPT-5-mini & 0.522 & 0.767 \\
\rowcolor{mementobg}
ReAct + GPT-5-mini + \textsc{Memento} & 0.525 \textcolor{deltagreen}{\scriptsize(+0.003)} & 0.825 \textcolor{deltagreen}{\scriptsize(+0.058)} \\
\midrule
\rowcolor{sectionbg}
\multicolumn{3}{l}{\textbf{Learning from web}} \\
\rowcolor{aetbg}
AET + Qwen & \textcolor{aettext}{0.552} \textcolor{deltagreen}{\scriptsize(+0.091)} & \textcolor{aettext}{0.767} \textcolor{deltagreen}{\scriptsize(+0.175)} \\
\rowcolor{bestbg}
\textsc{Memento} + Qwen & \textcolor{besttext}{0.579} \textcolor{deltagreen}{\scriptsize(+0.118)} & \textcolor{besttext}{0.808} \textcolor{deltagreen}{\scriptsize(+0.216)} \\
\rowcolor{aetbg}
AET + GPT-5-mini & \textcolor{aettext}{0.532} \textcolor{deltagreen}{\scriptsize(+0.010)} & \textcolor{aettext}{0.783} \textcolor{deltagreen}{\scriptsize(+0.016)} \\
\rowcolor{bestbg}
\textsc{Memento} + GPT-5-mini & \textcolor{besttext}{0.547} \textcolor{deltagreen}{\scriptsize(+0.025)} & \textcolor{besttext}{0.838} \textcolor{deltagreen}{\scriptsize(+0.071)} \\
\bottomrule
\end{tabular}
\caption{Framework performance on Sales Automation (Coverage Score) and Legal Research (Accuracy). \colorbox{mementobg}{\strut\small Green}: \textsc{Memento} memory augmentation. \colorbox{aetbg}{\strut\small Amber}: AET-only. \colorbox{bestbg}{\strut\small Blue}: full \textsc{Memento}. $\Delta$ in \textit{Learning from web} rows is relative to ReAct baselines w/o memory; others are relative to the row above.}
\label{tab:main-results}
\end{table}

\section{Results and Analysis}



\textbf{MEMENTO outperforms matched baselines across both tasks and
backbones.} On Sales Automation, MEMENTO + Qwen (0.579) improves over ReAct + Qwen (0.461) by $25.6\%$, while MEMENTO + GPT-5-mini (0.547) improves over its ReAct counterpart (0.522) by $4.8\%$. On Legal Research, MEMENTO + Qwen (0.808) improves over ReAct + Qwen (0.592) by $36.5\%$; with GPT-5-mini, the corresponding gain is $9.2\%$ (0.838 vs.\ 0.767). Table~\ref{tab:main-results} reports the full set of comparisons across parametric, in-context, and web-agent baselines. The consistency across an open-weight backbone and a hosted frontier backbone suggests that the gain is not tied to one underlying model. The gap does not stem from ReAct's inference budget: even with a $5\times$ larger step cap, ReAct self-terminates early and never uses the extra budget (Appendix~\ref{app:matched-compute}), indicating that the gap reflects how the agent searches rather than how much it is allowed to.

\textbf{Within-session exploration and cross-session memory make separable contributions.} Replacing flat ReAct with the AET, while withholding persistent memory raises Qwen's sales score from 0.461 to 0.552 ($+19.7\%$) and legal accuracy from 0.592 to 0.767 ($+29.6\%$). Adding memory on top of the same AET yields a further $4.9\%$ on sales and $5.3\%$ on legal. With GPT-5-mini, memory adds $2.8\%$ and $7.0\%$, respectively. Furthermore, ablations with cross-session memory showed that procedural stores drive the majority of the gain (roughly $85\%$). Details are provided in Appendix~\ref{app:ablation-details}.

\textbf{Improvements persist against stronger ReAct baselines.} We evaluated MEMENTO against three enhanced ReAct baselines (query rewriting, reranking, and their combination) on the sales task, with three independent judges (Qwen, GPT-4o, GPT-5.4). Table~\ref{tab:strong-baselines} depicts the results with Qwen as the judge. MEMENTO scores 0.447 versus 0.342 for the strongest enhanced ReAct baseline, a $30.7\%$ relative gain. Across four baselines and three judges, all 12 paired comparisons are statistically significant with bootstrap 95\% confidence intervals excluding zero (Appendix~\ref{app:strong-baselines}--\ref{app:judge-agreement}). Note that all five methods in Table~\ref{tab:strong-baselines} were run in the same time window against the same live web index, so absolute scores may slightly differ from scores in Table~\ref{tab:main-results}, but the within-run comparison is valid.


\begin{table}[!htbp]
\centering
\small
\begin{minipage}{0.58\linewidth}
\centering
\setlength{\tabcolsep}{5pt}
\begin{tabular}{lcc}
\toprule
\textbf{Method} & \textbf{Coverage} & \textbf{95\% CI} \\
\midrule
ReAct & 0.258 & [0.215, 0.302] \\
ReAct + Rerank & 0.295 & [0.239, 0.350] \\
ReAct + QR & 0.325 & [0.283, 0.366] \\
ReAct + QR + Rerank & 0.342 & [0.295, 0.391] \\
\textbf{MEMENTO} & \textbf{0.447} & \textbf{[0.408, 0.485]} \\
\bottomrule
\end{tabular}
\caption{MEMENTO vs.\ stronger ReAct baselines on the sales-task
re-run.}
\label{tab:strong-baselines}
\end{minipage}%
\hfill
\begin{minipage}{0.38\linewidth}
\centering
\setlength{\tabcolsep}{6pt}
\begin{tabular}{lc}
\toprule
\textbf{Method} & \textbf{Acc.} \\
\midrule
Closed-book Qwen & 0.414 \\
ReAct + Qwen & 0.439 \\
\textbf{MEMENTO + Qwen} & \textbf{0.561} \\
\bottomrule
\end{tabular}
\caption{Accuracy on 41 post-cutoff (2025–2026) Supreme Court cases.
Closed-book Qwen drops from 0.808 (original split), while MEMENTO
retains its lead on unexposed cases.}
\label{tab:recent-legal}
\end{minipage}
\end{table}

\textbf{A post-cutoff legal test set rules out memorization as an explanation for MEMENTO's gains.} On JUSTICE data, closed-book Qwen ($0.808$) outperforms every web-augmented variant (Table~\ref{tab:main-results}). The most plausible explanation is that the benchmark was seen during pretraining of the backbone model. To test whether MEMENTO's gains are separable from this exposure, we evaluate it on 41 Supreme Court cases from the 2025--2026 term that postdate the backbone Qwen's pretraining cutoff (construction in Section~\ref{sec:tasks-and-datasets}, Appendix~\ref{app:recent-legal}). On this unexposed set, closed-book Qwen's accuracy drops from $0.808$ to $0.414$, confirming the original number was inflated by memorization. MEMENTO reaches $0.561$ versus $0.439$ for ReAct (Table~\ref{tab:recent-legal}) on the new set, a $12.2$ point ($27.8\%$) improvement that provides direct evidence MEMENTO's advantage comes from web interaction, not memorized case outcomes. Details are provided in Appendix~\ref{app:recent-legal}.

\textbf{Gains generalize to a distributional prediction task.}
Subpopulation Opinion Prediction produces a full probability distribution
rather than free-form text (sales) or a class label (legal). On 100
SubPOP questions, MEMENTO reduces Total Variation Distance by $29.6\%$
and KL divergence by $46.8\%$ against the strongest baseline (ReAct with
query rewriting; Table~\ref{tab:subpop}, Appendix~\ref{app:subpop}).
Since SubPOP shares no output structure or evidence type with sales or
legal, the improvement suggests MEMENTO's AET and cross-session memory captures general-purpose research strategy (how to find and evaluate sources) rather than task-specific patterns.


\textbf{Training reduces inference cost while improving performance.} Under a fixed budget, adding memory to AET raises the score from 0.552 to 0.579 while reducing search queries by $6.6\%$, LLM calls by $4.1\%$, and repeated scraping by $26.0\%$ for sales task (Appendix~\ref{app:sales-test-details}). On legal research, memory reduces search queries by $19.8\%$, LLM calls by $19.3\%$, and wall-clock time by $28.0\%$ (Appendix~\ref{app:legal-efficiency}). Procedural memory drives $85\%$ of the memory-related gain (Appendix~\ref{app:ablation-details}). Furthermore, an audit of long-horizon memory behavior across six training batches shows stable rule support, minimal contradictions, and no cited domain on a known-misinformation index (Appendix~\ref{app:memory-audit}).

\section{Conclusion}
We propose treating the open web as a cross-session learning signal for task acquisition in low-data settings. MEMENTO combines adaptive within-session research with separately evolving procedural and declarative memory, allowing an agent to retain both what it learned and how it learned it without updating model weights. Across three
structurally distinct domains, MEMENTO improves over ReAct baselines
($+25.6\%$ on sales, $+36.5\%$ on legal research and $29.6\%$ TVD reduction on SubPOP opinion prediction), with gains that persist against stronger retrieval baselines (Table~\ref{tab:strong-baselines}) and on a post-cutoff Supreme Court test set constructed to rule out pretraining exposure ($+27.8\%$; Table~\ref{tab:recent-legal}). These results support a broader view of the web: not only as a source to retrieve from, but as an environment from which an agent can acquire transferable, auditable expertise.

\section*{AI Use Statement}

Generative AI tools were used to assist with coding-related tasks, including implementation support and debugging, as well as to improve the clarity and presentation of the manuscript. This included assistance with grammar, phrasing, sentence structure, and the organization and framing of written content. All AI-assisted outputs were reviewed and verified by the authors, who take full responsibility for the accuracy, content, claims, and final presentation of the work.


\bibliography{iclr2027_conference}
\bibliographystyle{iclr2027_conference}

\appendix

\section{Ethical Safeguards and Data Governance}
\label{app:ethics}

\paragraph{Dual-use risks.}
MEMENTO's capabilities may be misused for manipulative sales targeting or as a substitute for professional legal counsel. We therefore frame the system as decision support requiring human review rather than as an autonomous decision-maker. Our temporal cutoffs prevent the use of post-outcome evidence, while the human-readable memory enables periodic auditing of learned rules and sources (Appendix~\ref{app:memory-audit}). We intend to release the system under a responsible-use license prohibiting applications such as surveillance, doxxing, discriminatory profiling, and credit scoring.

\paragraph{Responsible web access.}
Our scraping service follows \texttt{robots.txt} directives and does not retrieve paths disallowed by a site's robots policy. Access is rate-limited using bounded semaphores permitting at most $8$ concurrent searches and $4$ concurrent page fetches. A persistent global URL cache prevents pages retrieved in previous runs from being fetched again. Deployments must additionally comply with site-specific terms of service. Learned procedural memory also reduces rather than amplifies web traffic: on the legal task, training reduces search queries by
$19.8\%$, LLM calls by $19.3\%$, and wall-clock time by $28.0\%$
(Appendix~\ref{app:legal-efficiency}); on the sales task, repeated URL
scraping falls by $26.0\%$ under the standard evaluation budget
(Appendix~\ref{app:sales-test-details}).

\paragraph{PII handling.}
The legal task is party-invariant: the predicted outcome should depend on the underlying facts and legal question, rather than the identities of the litigants. Party names and personal identifiers therefore provide no legitimate predictive signal and are removed from every legal-task input using the LLM-based PII-removal procedure described in Appendix~\ref{app:hyperparams}. In contrast, the sales task concerns the compatibility between a seller organization's capabilities and a prospective customer organization's needs. Organization identities are consequently task-defining information whose removal would make the task ill-posed. The sales inputs contain organizations i.e. the seller company, its products, and the prospective customer company rather than private individuals, and the ground-truth value propositions are derived from public customer-success stories published by the organizations themselves.

\paragraph{Data provenance.}
The experiments use publicly available, cited benchmarks and public organizational sources. Nevertheless, public availability does not guarantee accuracy or absence of bias. MEMENTO therefore preserves source information in its auditable memory, and outputs intended for high-stakes settings should be corroborated against independent sources and reviewed by qualified humans.

\section{Hyperparameters}
\label{app:hyperparams}

Both tasks share the same core configuration unless noted below. We train
on $60$ samples and evaluate on a held-out test set of $120$ samples per
task. The Adaptive Exploration Tree uses a \emph{question budget} of $25$
sub-questions per sample, of which $5$ are produced in the initial wave
and the remainder introduced over $2$ reflection checkpoints (yielding $3$
waves total). Each tool-augmented agent is capped at $10$ ReAct steps.
Cold-start exploration is enabled during training with a budget of $3$
search-and-scrape cycles, triggered whenever the relevance-filtered slice
of $M_1$ for the current sample is empty or sparse. The batch update loop
processes each training set in $6$ batches of size $10$. Actor, critique,
reflection, and synthesis calls run at temperature $0.7$. Unless stated
otherwise, all LLM calls and the LLM judge use Qwen-3.5-35B-A3B.

\paragraph{Sales Automation.}
Web search is restricted to pages published at least $6$ months before each
customer-success story's publication date. At inference time the exploration
budget is set to $0$: memory is consumed but not extended. The task inputs are
organizations (seller, products, and prospective customer), not private
individuals. Entity names are therefore retained because removing them would
erase the task-defining signal.

\paragraph{Legal Research.}
Web search uses a $5$-year ($60$-month) look-back
window relative to each case's decision date, ensuring the agent cannot
retrieve the ruling itself. To prevent parametric leakage of the case
outcome, every input is passed through an LLM-based PII-removal step that
strips party names and personal identifiers, and an explicit anti-cheat
instruction is injected into the ReAct system prompt forbidding reliance
on internal training knowledge of the case's disposition. Unlike the sales
task, the exploration budget at inference time is held at $3$ so that
under-trained legal domains can still gather missing context, though no
artifacts are written back to long-term memory.

\section{Compute Resources}
\label{sec:compute}

All Qwen-3.5-35B-A3B calls are served from a self-hosted SGLang deployment running across $4$ nodes of $8\times$NVIDIA A100 80\,GB GPUs ($32$ GPUs total), exposed through a least-inflight load-balancer. The GPT-5-mini variants are accessed through Azure OpenAI's hosted endpoint and incur no local GPU compute. Web search and HTML scraping are handled by an in-house Crawl4AI-based service shared across all experiments. The crawler follows \texttt{robots.txt} directives and does not fetch disallowed paths. Concurrent access is gated by two bounded semaphores ($8$ simultaneous \textsc{Search\_Web} queries and $4$ simultaneous \textsc{Scrape\_Results} fetches), while a persistent global URL cache prevents re-fetching pages already retrieved in prior runs. These controls bound request bursts and reduce repeated load; deployments must additionally comply with site-specific terms of service.

All runs use a thread pool of $20$--$30$ concurrent workers. End-to-end wall times for the sales task are summarized in Table~\ref{tab:compute}. A $60$-sample training run takes approximately $6$ hours with Qwen and $7$ hours with GPT-5-mini; prediction over $120$ samples takes approximately $6$ and $8$ hours, respectively. Despite hosted inference, end-to-end latency remains dominated by search, scraping, and API round-trips under concurrent load. Because all experiments share the same web-service quota, wall-clock time also depends on how many runs are active simultaneously.

\begin{table}[h]
\centering
\small
\begin{tabular}{l l l l}
\toprule
\textbf{Stage} & \textbf{Backend} & \textbf{Compute} & \textbf{Wall time} \\
\midrule
Training (60 samples)    & Qwen-3.5-35B-A3B & $4 \times 8\times$A100 80\,GB & 6 h \\
Training (60 samples)    & gpt-5-mini (Azure) & API only                       & $\sim$7 h \\
Prediction (120 samples) & Qwen-3.5-35B-A3B & $4 \times 8\times$A100 80\,GB & 6 h \\
Prediction (120 samples) & gpt-5-mini (Azure) & API only                       & $\sim$8 h \\
\bottomrule
\end{tabular}
\caption{Compute resources and wall-clock times for the sales-task experiments.}
\label{tab:compute}
\end{table}


\section{Additional Sales Automation Results}
\label{app:sales-test-details}

\begin{table}[H]
\centering
\small
\setlength{\tabcolsep}{5pt}
\resizebox{\linewidth}{!}{
\begin{tabular}{l cc cc}
\toprule
& \multicolumn{2}{c}{\texttt{max\_agent\_steps = 20} (a)} & \multicolumn{2}{c}{\texttt{max\_agent\_steps = 10} (b)} \\
\cmidrule(lr){2-3} \cmidrule(lr){4-5}
\textbf{Metric} & After training & Without training & After training & Without training \\
\midrule
\textbf{Score} & \textbf{0.5864} & 0.5691 & \textbf{0.5791} & 0.5526 \\
\midrule
\multicolumn{5}{l}{\textit{Decomposition behaviour}} \\
Avg.\ Questions / Sample      & 21.18    & 22.78    & 22.13    & 23.52 \\
Avg.\ Agent Steps / Question  & 9.80     & 10.93    & 7.80     & 8.71 \\
Avg.\ Scratchpad Size (chars) & 40{,}122 & 48{,}475 & 41{,}172 & 39{,}188 \\
\midrule
\multicolumn{5}{l}{\textit{Search discipline}} \\
Avg.\ Search Queries / Sample & 174.00 & 201.15 & 139.61 & 149.49 \\
Avg.\ Unique Google URLs      & 786.84 & 804.88 & 659.13 & 644.75 \\
Avg.\ Repeated Google URLs    & 539.58 & 698.84 & 410.97 & 500.71 \\
\midrule
\multicolumn{5}{l}{\textit{Scraping behaviour}} \\
Avg.\ Unique URLs Scraped   & 47.76 & 47.04 & 37.38 & 36.55 \\
Avg.\ Repeated URLs Scraped & 28.96 & 37.08 & 19.02 & 25.69 \\
\midrule
\multicolumn{5}{l}{\textit{Compute}} \\
Avg.\ LLM Calls & 398.21 & 441.01 & 310.79 & 323.97 \\
\bottomrule
\end{tabular}}
\caption{Test-set performance on 120 held-out samples, grouped by the pipeline stage each metric reflects.}
\label{tab:test-results}
\end{table}

\subsection{Exploration Phase}

To bootstrap $M_1$ before any supervised signal is available, the system runs a \textit{pre-solve exploration phase} when craft knowledge is absent from memory (left panel of Figure~\ref{fig:pipeline}). At the start of solving a sample, if $M_1$ is empty, an exploration agent issues a fixed number of search-and-scrape cycles to discover domain strategies and methodologies, populating $M_1$ with a foundational procedural prior before inference begins. If $M_1$ already contains knowledge, this step is skipped.

Additionally, at the end of solving each sample, if exploration budget remains and the sample produced a valid answer, the agent re-enters exploration to further enrich $M_1$ based on what was encountered during that run.

\section{Ablation: Procedural vs.\ Declarative Memory}
\label{app:ablation-details}

To isolate the contribution of each memory channel, we evaluate four
configurations on the 120 held-out sales samples with a budget of 10
agent steps per question: no memory (\emph{neither}), declarative only
($M_4$), procedural only ($M_1, M_2, M_3$), and the full system.
Table~\ref{tab:ablation-score} reports the results.

\begin{table}[H]
\centering
\small
\setlength{\tabcolsep}{6pt}
\begin{tabular}{l cc}
\toprule
\textbf{Configuration} & \textbf{Score} & $\Delta$ vs.\ neither \\
\midrule
Neither (zero-shot)               & 0.5526 & --- \\
Declarative only ($M_4$)          & 0.5533 & $+0.0007$ \\
Procedural only ($M_1, M_2, M_3$) & 0.5751 & $+0.0225$ \\
\textbf{Both (full system)}       & \textbf{0.5791} & $\mathbf{+0.0265}$ \\
\bottomrule
\end{tabular}
\vspace{4pt}
\caption{Ablation scores on 120 held-out sales samples at \texttt{max\_agent\_steps\,=\,10}.}
\label{tab:ablation-score}
\end{table}

\paragraph{Procedural memory accounts for the majority of the gain.}
Of the $+0.0265$ total improvement, procedural memory alone delivers
$+0.0225$ --- roughly $85\%$ of the lift. Declarative memory contributes
only $+0.0007$ in isolation, within the range of run-to-run noise. The
implication is clear: for this task, \emph{how} the agent searches ---
query formulation heuristics, decomposition rules, known dead ends ---
matters far more than \emph{what} entity-level facts it has pre-cached.

This finding aligns with the cognitive motivation of our architecture.
In ACT-R, novice-to-expert progression is driven primarily by production
compilation: the conversion of slow, deliberate strategies into efficient
automatic routines~\cite{anderson2004integrated}. Declarative knowledge
enables reasoning but does not, on its own, make the reasoner faster or
more directed. Our ablation provides an empirical parallel: the procedural
stores ($M_1, M_2, M_3$) reshape the agent's search trajectory, while
the declarative store ($M_4$) leaves it largely unchanged.

\paragraph{Procedural memory improves quality and reduces cost simultaneously.}
Beyond the score gain, the procedural variants are uniformly more
compute-efficient. Compared to the no-memory baseline, procedural-only
reduces search queries by $5.9\%$, LLM calls by $4.5\%$, and ---
most strikingly --- repeated URL scraping by $24.3\%$
($25.69 \rightarrow 19.45$). The learned search rules and failure
patterns are not merely guiding the agent toward better evidence;
they are actively pruning redundant exploration.

\paragraph{Declarative memory alone is counterproductive.}
The declarative-only configuration is the most expensive of all four
settings: it generates the most questions per sample ($24.61$ vs.\
$23.52$ for neither), the largest scratchpad ($46{,}917$ chars), and
the most LLM calls ($328.34$), yet achieves only a negligible score
improvement. Without procedural guidance to direct search, cached
facts appear to encourage the agent to pursue more leads rather than
better ones, inflating the research trajectory without compounding
into stronger outputs. Full per-configuration behavioural metrics
are reported in Table~\ref{tab:ablation-full}.

\paragraph{The two channels are weakly complementary.}
If the effects were strictly independent, the combined lift would be
approximately $0.0225 + 0.0007 = 0.0232$. The full system achieves
$+0.0265$, a small super-additive effect of $\approx 0.003$. This is
consistent with the interpretation that declarative facts become useful
\emph{conditional on} good procedural strategy: once the agent knows
how to search and which paths to avoid, having entity-specific facts
on hand allows it to short-circuit a few lookups that would otherwise
be redundant. The behavioural metrics for the full system are
essentially indistinguishable from procedural-only --- the score
moves, the trajectory does not.

\begin{table}[H]
\centering
\small
\setlength{\tabcolsep}{4pt}
\begin{tabular}{l cccc}
\toprule
\textbf{Metric} & Procedural & Both & Declarative & Neither \\
Score                       & 0.5751   & \textbf{0.5791} & 0.5533   & 0.5526 \\
Avg.\ questions / sample    & 21.95    & 22.13           & 24.61    & 23.52 \\
Avg.\ agent steps / question& 7.82     & 7.61            & 7.72     & 8.71 \\
Avg.\ scratchpad (chars)    & 42{,}059 & 41{,}172        & 46{,}917 & 39{,}188 \\
Avg.\ search queries        & 140.63   & 139.61          & 145.91   & 149.49 \\
Avg.\ unique URLs scraped   & 38.42    & 37.38           & 41.52    & 36.55 \\
Avg.\ repeated URLs scraped & 19.45    & 19.02           & 26.72    & 25.69 \\
Avg.\ repeated Google URLs  & 407.10   & 410.97          & 461.56   & 500.71 \\
Avg.\ LLM calls             & 309.28   & 310.79          & 328.34   & 323.97 \\
\bottomrule
\end{tabular}
\caption{Per-configuration behavioural metrics for the four ablation settings.}
\label{tab:ablation-full}
\end{table}

\section{Is the Gain a Compute Artifact?}
\label{app:matched-compute}

A natural concern is that MEMENTO's advantage over ReAct stems from its larger
inference budget rather than its search architecture. To test this, we grant the
ReAct baseline a generous cap of 50 steps, where each step is a single
\textsc{Search\_Web}, \textsc{Scrape\_Results}, or \textsc{Generate\_Answer}
action and the agent self-terminates once it produces a final answer.

Even with this five-fold larger budget, ReAct self-terminates at a median of
${\sim}10$ steps and never exceeds 25; the most search-intensive sample uses only
25 of the 50 available steps. Because the cap is never binding, enlarging it
cannot change ReAct's outputs---the additional budget is simply never exercised.

The ReAct--MEMENTO gap is therefore not a budget-starvation effect. ReAct does
not underperform because it runs out of steps, but because it has no mechanism to
recognize that further search would help before it stops. MEMENTO's Adaptive
Exploration Tree instead reflects after each wave, identifies remaining evidence
gaps, and deliberately continues searching before synthesis. The advantage is a
difference in \emph{when} and \emph{why} to keep searching, not in how much
compute is nominally available.

\section{Sales Automation Evaluation Rubric}
\label{app:rubric}

This appendix details the LLM-judge evaluation protocol used for the Sales Automation task referenced in Section~4.3. All judging is performed with Qwen-3.5-35B-A3B using a single prompt per (ground-truth point, candidate pitch) pair.

\paragraph{Judge persona.} The judge is instructed with the following persona:
\begin{quote}
\textit{``You are a Senior Sales Enablement Evaluator. You are grading on Sales Effectiveness and Factual Precision.''}
\end{quote}

\paragraph{Scoring procedure.} For each ground-truth winning pitch point, the judge selects the best-matching candidate point from the system's prediction and assigns an integer score on a 0--5 scale (Table~\ref{tab:sales-rubric}). The per-sample \emph{normalized score} is then computed as
\[
\text{normalized\_score}
=
\frac{1}{5 \lvert G \rvert}
\sum_{g \in G} s(g)
\]
where $G$ is the set of ground-truth pitch points for the sample and $s(g) \in \{0,1,2,3,4,5\}$ is the score assigned to point $g$. The reported \emph{average score} in Section~5 is the mean of normalized\_score across the test set.

\begin{table}[H]
\centering
\small
\setlength{\tabcolsep}{6pt}
\renewcommand{\arraystretch}{1.25}
\begin{tabular}{c l p{8.6cm}}
\toprule
\textbf{Score} & \textbf{Label} & \textbf{Description} \\
\midrule
0 & Miss / Irrelevant & Candidate completely misses this concept. No mention of the feature, benefit, or metric. \\
1 & Marketing Fluff & Vaguely mentions the topic (e.g.\ ``improved efficiency'') but lacks any specific substance. Generic platitude. \\
2 & Topic Match & Identifies the correct product or pain point, but misses the specific solution or outcome. E.g.\ GT says ``reduced downtime by 40\%,'' candidate says ``helps with downtime.'' \\
3 & Implied / Soft Match & Captures the core value proposition correctly, but misses \emph{hero evidence} (specific numbers, names, unique mechanisms). Solid conversational point but less persuasive than GT. \\
4 & Strong Sales Argument & Captures core value \emph{and} the key mechanism/outcome. Persuasive and accurate. May miss a minor detail (date, city) that doesn't affect persuasion. \\
5 & Strategic Bullseye & Perfect extraction: product + pain + value + specific metric/evidence, essentially verbatim from GT. ``This is exactly why they bought.'' \\
\bottomrule
\end{tabular}
\caption{Rubric used by the LLM judge to score each ground-truth pitch point against the candidate pitch.}
\label{tab:sales-rubric}
\end{table}

\section{Generalization to Held-Out Samples}

We now evaluate whether the gains observed during training transfer to unseen samples. Table~\ref{tab:test-results} reports performance on the 120-sample test set, comparing the trained system against a cold-start ablation under two per-question compute budgets: a generous budget of 20 agent steps per sub-question (run~(a)) and a tighter budget of 10 (run~(b)).

To make the effect of training easier to read, Table~\ref{tab:test-deltas} recasts the same numbers as relative changes induced by training, separately for each budget.

\paragraph{Finding: learned discipline wins under a tight budget and generalizes beyond training.}
Under the more realistic per-question budget of 10 agent steps (run~(b)), training raises the score from 0.5526 to 0.5791 (a $+4.8\%$ relative gain) while simultaneously reducing search queries by $6.6\%$, LLM calls by $4.1\%$, agent steps per question by $10.4\%$, and repeated URL scraping by $26\%$. The same tightening appears one layer deeper in the pipeline: repeated Google URLs returned by the search API drop by $17.9\%$. Importantly, these gains do not reflect overfitting: there is no overlap in triples between train and test, the learned procedural efficiency patterns transfer to unseen samples, and the improvements persist across different compute budgets.

\begin{table}[H]
\centering
\small
\setlength{\tabcolsep}{6pt}
\begin{tabular}{l cc}
\toprule
\textbf{Metric} & \textbf{Run (a), budget $=20$} & \textbf{Run (b), budget $=10$} \\
\midrule
Score                  & $+3.0\%$ (better) & $+4.8\%$ (better) \\
Search queries         & $-13.5\%$         & $-6.6\%$ \\
Repeated Google URLs   & $-22.8\%$         & $-17.9\%$ \\
Repeated URLs scraped  & $-21.9\%$         & $-26.0\%$ \\
Unique URLs scraped    & $+1.5\%$          & $+2.3\%$ \\
LLM calls              & $-9.7\%$          & $-4.1\%$ \\
Agent steps / question & $-10.3\%$         & $-10.4\%$ \\
\bottomrule
\end{tabular}
\caption{Relative change from \textit{no training} to \textit{after training}. Positive values indicate increase; for Score, positive is better. Efficiency metrics improve under training in both runs.}
\label{tab:test-deltas}
\end{table}

\section{Statistical Significance of the Sales Task Scores}
\label{app:sales-ci}

We report 95\% confidence intervals (CI) for the mean Sales Task scores using a non-parametric bootstrap procedure. For each method, we repeatedly resample the per-sample scores with replacement, compute the mean score for each resample, and take the 2.5th and 97.5th percentiles of the resulting distribution of means as the confidence interval bounds. We use 1000 bootstrap iterations. This approach avoids assuming normally distributed scores and directly estimates the sampling variability from the observed data.

\begin{table}[ht]
\centering
\begin{tabular}{llc}
\hline
\textbf{Approach family} & \textbf{Configuration} & \textbf{95\% CI} \\
\hline
Closed-book & Qwen                 & [0.386, 0.446] \\
           & Qwen + Memory        & [0.402, 0.469] \\
           & GPT-5-mini           & [0.339, 0.445] \\
           & GPT-5-mini + Memory  & [0.345, 0.455] \\
\hline
In-context & 5-shot Qwen          & [0.343, 0.408] \\
           & 5-shot Qwen + Memory & [0.348, 0.415] \\
           & 5-shot GPT           & [0.417, 0.507] \\
           & 5-shot GPT + Memory  & [0.429, 0.519] \\
\hline
Web agent & ReAct + Qwen          & [0.428, 0.495] \\
           & ReAct + Qwen + Memory & [0.435, 0.502] \\
           & ReAct + GPT           & [0.486, 0.556] \\
           & ReAct + GPT + Memory  & [0.491, 0.559] \\
\hline
Learning from web & Qwen + AET            & [0.515, 0.590] \\
           & Qwen + AET + MEMENTO  & [0.541, 0.614] \\
           & GPT + AET             & [0.500, 0.565] \\
           & GPT + AET + MEMENTO   & [0.516, 0.583] \\
\hline
\end{tabular}
\caption{95\% bootstrap confidence intervals for the Sales Task score.}
\label{tab:sales-ci}
\end{table}

The bootstrap confidence interval is computed as follows. Given per-sample scores
$s = [s_1, s_2, \ldots, s_N]$, we repeatedly sample $N$ scores with replacement,
compute the resampled mean, and repeat this process 1000 times to obtain a distribution
of bootstrap means. The 95\% CI is then given by the 2.5th and 97.5th percentiles of
this distribution.

\begin{verbatim}
def bootstrap_ci(scores, n_iter=1000):
    boot = [np.random.choice(scores,
                             size=len(scores),
                             replace=True).mean()
            for _ in range(n_iter)]
    lo = np.percentile(boot, 2.5)
    hi = np.percentile(boot, 97.5)
    return scores.mean(), lo, hi
\end{verbatim}


\section{Legal Task: Training Reduces Cost and Improves Efficiency}
\label{app:legal-efficiency}

We evaluate the trained system on a 120-sample legal-task benchmark.
Table~\ref{tab:legal-training-comparison}
contrasts the baseline agent (\emph{no training}) with the fully trained
agent (\emph{with training}); the latter receives the procedural and declarative
artifacts learned during training.

Across all efficiency metrics, training yields consistent improvements.
The trained agent requires fewer questions per sample (20.61 vs.\ 23.68),
fewer agent steps per question (6.63 vs.\ 7.29), and substantially fewer
search queries (116.09 vs.\ 144.72). It also reduces total web interaction,
including both unique and repeated URL fetches, as well as Google result
exploration.

Overall, training significantly improves system efficiency, reducing
both latency and resource usage while maintaining a similar problem-solving
trajectory.

\begin{table}[H]
\centering
\small
\setlength{\tabcolsep}{6pt}
\begin{tabular}{l cc r}
\toprule
\textbf{Metric} & \textbf{Baseline} & \textbf{Trained} & $\Delta$ \\
\midrule
Total samples                            & 120     & 120     & --- \\
\midrule
Avg.\ questions per sample               & 23.68   & 20.61   & $-13.0\%$ \\
Avg.\ agent steps per question           & 7.29    & 6.63    & $-9.1\%$ \\
Avg.\ search queries per sample          & 144.72  & 116.09  & $-19.8\%$ \\
Avg.\ unique URLs scraped                & 16.12   & 15.77   & $-2.2\%$ \\
Avg.\ repeated URLs scraped              & 13.69   & 11.41   & $-16.7\%$ \\
Avg.\ total URLs scraped                 & 29.81   & 27.18   & $-8.8\%$ \\
Avg.\ unique Google URLs                 & 420.39  & 327.13  & $-22.2\%$ \\
Avg.\ repeated Google URLs               & 581.80  & 481.99  & $-17.2\%$ \\
Avg.\ short-term scratchpad size (chars) & 37684.12& 32738.99& $-13.1\%$ \\
Avg.\ time per sample (s)                & 4769.92 & 3436.06 & $-28.0\%$ \\
Avg.\ LLM calls                          & 295.01  & 238.01  & $-19.3\%$ \\
\bottomrule
\end{tabular}
\caption{Trained vs.\ baseline agent on the legal task. Metrics computed over $120$ samples.}
\label{tab:legal-training-comparison}
\end{table}

\paragraph{Interpretation.}
The pattern matches what we observed in the SDR ablations
(Table~\ref{tab:ablation-full}): procedural artifacts -- search rules,
craft knowledge, and failure patterns -- shape the trajectory toward
shorter, less redundant exploration. The agent asks fewer questions,
re-scrapes fewer pages, and writes a tighter scratchpad, while still
producing slightly better final answers. In other words, training does
not simply trade compute for accuracy; it improves both axes simultaneously.
For high-baseline tasks like this one, the dominant value of training is
\emph{efficiency}, with accuracy gain a useful by-product.

\section{Stronger Web-Agent Baselines}
\label{app:strong-baselines}

To test whether MEMENTO's advantage comes only from better query construction
or source ordering, we re-ran the sales evaluation with three enhanced ReAct
baselines: query rewriting (QR), reranking, and their combination. All five
methods were run in the same time window against the same live index and were
scored by the same judges. Because the web index changes over time, these
absolute scores are not expected to match the earlier runs in
Table~\ref{tab:main-results}; the controlled comparison is within this re-run.

We additionally perform paired non-parametric bootstrap tests on per-sample
differences and Wilcoxon signed-rank tests. Table~\ref{tab:paired-significance}
reports the paired effect estimates. The paired analysis uses the matched set
of valid outputs for each comparison, so its estimate can differ slightly
from subtracting the rounded all-output means in
Table~\ref{tab:strong-baselines}. Every confidence interval excludes zero;
the corresponding Wilcoxon $p$-values range from $1.1\times10^{-3}$ to
$1.8\times10^{-11}$.

\begin{table}[H]
\centering
\scriptsize
\setlength{\tabcolsep}{3.2pt}
\resizebox{\linewidth}{!}{
\begin{tabular}{lccc}
\toprule
\textbf{Comparison} & \textbf{Qwen: $\Delta$ (95\% CI)} & \textbf{GPT-4o: $\Delta$ (95\% CI)} & \textbf{GPT-5.4: $\Delta$ (95\% CI)} \\
\midrule
MEMENTO vs.\ ReAct & +0.212 [0.159, 0.267] & +0.267 [0.212, 0.321] & +0.182 [0.141, 0.223] \\
MEMENTO vs.\ ReAct+QR & +0.134 [0.078, 0.192] & +0.197 [0.141, 0.255] & +0.118 [0.073, 0.163] \\
MEMENTO vs.\ ReAct+rerank & +0.141 [0.066, 0.215] & +0.196 [0.123, 0.268] & +0.140 [0.078, 0.200] \\
MEMENTO vs.\ ReAct+QR+rerank & +0.094 [0.037, 0.149] & +0.167 [0.104, 0.227] & +0.102 [0.055, 0.148] \\
\bottomrule
\end{tabular}}
\caption{Paired coverage improvements across three independent judges.}
\label{tab:paired-significance}
\end{table}

\section{Inter-Judge Agreement}
\label{app:judge-agreement}

We validate the sales-task evaluator by scoring every method--sample pair with
Qwen, GPT-4o, and GPT-5.4. Across the 600 method--sample aggregates, pairwise
Pearson correlations range from 0.874 to 0.899 and Spearman correlations from
0.863 to 0.892. The absolute-agreement intraclass correlations are
ICC$(2,1)=0.868$ for a single judge and ICC$(2,k)=0.952$ for the three-judge
average. Agreement does not cluster by model family: weighted agreement is
$\kappa=0.821$ for Qwen/GPT-4o, $0.868$ for Qwen/GPT-5.4, and $0.823$ for
GPT-4o/GPT-5.4. We therefore use the three-judge average for the controlled
re-run and retain the original single-judge results for comparability with
the initial experiments.

\section{Post-Cutoff Legal Evaluation}
\label{app:recent-legal}

Table~\ref{tab:recent-legal} in the main text reports accuracy on 41 U.S. Supreme Court cases decided in the 2025--2026 term. The original JUSTICE~\citep{alali2021justice} cases were compiled by 2021 and may have appeared in pretraining corpora. We therefore construct an additional set of 41 U.S. Supreme Court cases decided in the 2025--2026 term from Oyez, following the same input formatting and party-removal procedure as JUSTICE. These cases postdate both the benchmark compilation and the backbone Qwen model's pretraining window, making direct parametric exposure substantially less plausible. As with the original JUSTICE evaluation, web search on this set is restricted to sources published at least 60 months before each case's decision date.

Closed-book Qwen falls from 0.808 on the original test split (Table~\ref{tab:main-results}) to 0.414 (Table~\ref{tab:recent-legal}) on the post-cutoff cases, confirming that the original closed-book number is substantially affected by benchmark exposure while the new test set is unexposed and suitable for evaluation. On the new test set, MEMENTO improves over the ReAct baseline by 12.2 accuracy points ($+27.8\%$) indicating it is able to leverage web through AET and dual channel memory in an improved manner. MEMENTO correctly answered 5 cases that ReAct missed, while ReAct did not correctly answer any cases that MEMENTO missed. This difference was not statistically significant under an exact two-sided McNemar test ($p = 0.063$), although the observed difference was entirely in favor of MEMENTO.

\section{Third Domain: Subpopulation Opinion Prediction}
\label{app:subpop}

\paragraph{Motivation.}
Surveys are the primary instrument for measuring public opinion, but they are
expensive and slow to field. A system that anticipates how a demographic
subgroup will answer a question supports questionnaire design and reasoning
about under-sampled groups. The task also matches the low-data setting studied
in this paper: subgroup-level ground truth exists only for questions that have
already been fielded, while the evidence bearing on a question (related polls,
demographic research, issue coverage) is spread across heterogeneous web
sources. Answering well therefore depends on knowing where authoritative
demographic evidence lives and how to combine signals about a subgroup's
composition. We expect this kind of procedural knowledge to transfer across
questions.

\paragraph{Task and data.}
SubPOP~\citep{subpop2025} is drawn from Pew Research Center's American Trends
Panel~\citep{pew_atp} and NORC's General Social Survey~\citep{gss2022}. Each item pairs a survey question with
a target demographic subgroup; the system receives the question, its fixed
answer options, and the subgroup, and outputs a probability distribution over
the options. We evaluate on 100 questions spanning eight demographic
attributes: region, education, income, ideology, political party, race,
religion, and sex. Unlike sales (free-form generation) and legal research
(binary classification), the output is scored as a full distribution against
the observed subgroup distribution, using TVD and $\mathrm{KL}(q\,\|\,p)$ as
defined in Section~\ref{sec:subpop-metrics}.

\paragraph{Metrics.}
Let $p$ be the predicted distribution over answer options and $q$ the observed subgroup distribution. We report Total Variation Distance, $\mathrm{TVD}(p,q)=\frac{1}{2}\sum_i |p_i-q_i|$, a bounded measure in $[0,1]$ of misallocated probability mass, and KL divergence, $\mathrm{KL}(q\,\|\,p)=\sum_i q_i\log(q_i/p_i)$, which more heavily penalizes assigning low probability to options the subgroup actually chooses. Lower is better for both.

\paragraph{Setup.}
All methods use the Qwen backbone and date-restricted web search, following the protocol of the other two tasks. We compare against three baselines: the LLM without web access, ReAct with web search, and ReAct with query rewriting and web search.

\begin{table}[H]
\centering
\small
\begin{tabular}{lcc}
\toprule
\textbf{Method} & \textbf{TVD $\downarrow$} & \textbf{KL $\downarrow$} \\
\midrule
\textbf{MEMENTO} & \textbf{0.2843} & \textbf{0.2742} \\
ReAct + web & 0.4115 & 0.5537 \\
ReAct + QR + web & 0.4036 & 0.5154 \\
LLM without web & 0.4498 & 0.6369 \\
\bottomrule
\end{tabular}
\caption{SubPOP results on 100 questions. Lower is better.}
\label{tab:subpop}
\end{table}

\paragraph{Results.}
Web access helps on its own: ReAct reduces TVD by 8.5\% and KL by 13.1\%
relative to the LLM without web. Query rewriting adds little on top (a further
1.9\% on TVD and 6.9\% on KL). MEMENTO reduces TVD by 29.6\% and KL by 46.8\%
relative to the strongest baseline (ReAct + QR + web), a larger step than web
access and query rewriting combined. The gain is larger on KL, which penalizes
near-zero probability on answers a subgroup actually gives. This suggests that
the baselines more often commit to confident but wrong distributions.

We hypothesize that the gain comes from procedural memory that accumulates
transferable strategies for locating authoritative demographic sources and
reasoning about subgroup composition, so that each estimate starts from better
evidence than a baseline searching from scratch.

\section{Long-Horizon Memory Behavior and Source Audit}
\label{app:memory-audit}

We analyze the six sales-task training batches along three dimensions:
whether learned rules remain supported, whether memory grows without bound,
and whether the web evidence entering memory is credible.

\paragraph{Rule stability.}
Every web-action rule records independent batch-level support and
contradiction events. Table~\ref{tab:rule-stability} shows that the catalog
grows from 10 to 29 rules while total support rises from 10 to 105. Only two
contradiction events occur, both against one minor heuristic limiting repeated
PDF-scraping attempts; it is down-weighted once contradictions exceed
support. No rule is retracted during this six-batch window.

\begin{table}[H]
\centering
\small
\begin{tabular}{ccccc}
\toprule
\textbf{Batch} & \textbf{\# rules} & \textbf{Total support} & \textbf{Contradictions} & \textbf{Rules contradicted} \\
\midrule
1 & 10 & 10  & 0 & 0 \\
2 & 13 & 23  & 0 & 0 \\
3 & 18 & 41  & 0 & 0 \\
4 & 24 & 61  & 1 & 1 \\
5 & 26 & 82  & 2 & 1 \\
6 & 29 & 105 & 2 & 1 \\
\bottomrule
\end{tabular}
\caption{Evolution of procedural rules over six training batches.}
\label{tab:rule-stability}
\end{table}

\paragraph{Memory size and revision.}
The combined procedural memory grows from 52,203 to 151,913 characters, but
not through append-only accumulation. Every batch modifies existing entries;
batches 3 and 6 delete substantial material, and batch 6 ends with a net
shrink as consolidation removes more text than it adds.

\begin{table}[H]
\centering
\small
\begin{tabular}{crrrr}
\toprule
\textbf{Batch} & \textbf{Chars after} & \textbf{Added} & \textbf{Deleted} & \textbf{Modified} \\
\midrule
1 & 52,203  & 52,203 & 0      & 0 \\
2 & 55,075  & 284    & 0      & 29,292 \\
3 & 78,595  & 12,293 & 3,680  & 15,111 \\
4 & 79,564  & 409    & 633    & 24,500 \\
5 & 153,046 & 16,150 & 0      & 57,984 \\
6 & 151,913 & 1,340  & 12,937 & 21,119 \\
\bottomrule
\end{tabular}
\caption{Memory editing activity. ``Modified'' counts characters participating in rewrites and is not a net-change column.}
\label{tab:memory-size}
\end{table}

\paragraph{Source quality.}
We audit every page cited across the 60 sales training tasks (approximately
700 pages from 400 domains) against the Tranco top-sites ranking
~\citep{tranco2019} and the 2,674-domain snapshot of the Iffy Index
~\citep{iffyindex}. No cited domain appears on the misinformation
list. Approximately 70\% of sources are clearly credible, 8\% are moderate,
and the remaining 22\% are legitimate specialist sources such as
manufacturers, trade journals, and industry associations. The clearly
credible share remains approximately 70--77\% across all six batches. This
audit does not prove factual correctness, but it provides evidence that the
source mix does not degrade as memory accumulates.

\section{Representative Memory Entries}
\label{app:memory-examples}

\begin{description}[leftmargin=3.5cm,labelwidth=3.1cm,style=multiline,nosep]
\item[$M_1$ Craft knowledge.] Use revenue-per-employee triangulation to estimate a private company's finances when direct data is unavailable. When to use: prospecting private companies whose revenue isn't publicly disclosed. Why it works: public data rarely reveals financial health directly; industry-specific per-employee multipliers create reliable tiering assumptions for sales-fit scoring.
\item[$M_2$ Decomposition rule.] Each root child must be self-contained and understandable in isolation. Explicitly name the product, entity, and evidence target inside each question whenever relevant.
\item[$M_3$ Web-action rule.] [query] Formulate queries using concise natural language and avoid keyword stuffing. [urlselection] Prioritize HTML pages over PDFs for initial content extraction.
\item[$M_4$ Domain fact.] Seller ABC: Positions itself as an Intelligent Agreement Management platform covering the full agreement lifecycle; offers a 'Signature Appliance' for on-premises, air-gapped, highly-regulated deployments. Product XYZ: Supports cloud-native architecture (Strata) for faster deployment than legacy monolithic systems; can integrate AI/OCR/NLP to process unstructured data like accident forms.
\end{description}

Entries are filtered, canonicalized, and contradiction-weighted. Deployment should add independent-source corroboration and periodic human review.

\section{Limitations and Broader Impact}
\label{app:limitations}

\paragraph{Limitations and Future Work.}
MEMENTO addresses low \emph{labeled-data} availability, not low compute: the AET is more expensive during training than linear retrieval and depends on substantial LLM and web-search infrastructure. Absolute results can vary as live search indexes change, although all methods in each comparison are run in the same time window. Sales evaluation still
relies on LLM judges despite high three-judge agreement (Appendix~\ref{app:judge-agreement}). The long-horizon audit covers six
batches, so drift over months of online use remains unknown.
Performance depends on web coverage and source quality, and
generalization beyond the three evaluated domains remains open. Memory
is frozen at test time, preventing continual adaptation but also
limiting uncontrolled post-deployment drift. Future work should study
longer horizons, human evaluation, frozen web snapshots, online updates
with rollback, and broader domains and languages.

\paragraph{Broader Impact.}
MEMENTO's human-readable memory improves auditability relative to
weight-level adaptation, but the same capability could enable
surveillance, doxxing, discriminatory profiling, manipulative
targeting, or automated opposition research. In high-stakes domains,
outputs should augment rather than replace professional judgment.
Before public deployment, memory writes should require corroboration
across independent sources and periodic human audit. The system also
inherits demographic, linguistic, and misinformation biases from the
web.

\end{document}